\documentclass[sigconf]{acmart}
\usepackage{multirow}
\usepackage{makecell}
\newcolumntype{C}[1]{>{\centering\arraybackslash}p{#1}}

\usepackage{enumitem}
\AtBeginDocument{%
  }

\copyrightyear{2026}
\acmYear{2026}
\setcopyright{cc}
\setcctype{by-nc-nd}
\acmConference[CIKM '26]{Proceedings of the 35th ACM International Conference on Information and Knowledge Management}{November 07--11, 2026}{Rome, Italy}
\acmBooktitle{Proceedings of the 35th ACM International Conference on Information and Knowledge Management (CIKM '26), November 07--11, 2026, Rome, Italy}
\acmDOI{10.1145/3799682.3840763}
\acmISBN{979-8-4007-2539-5/2026/11}

\begin{document}

\title{Evidence-Consistent Generative Detection under Scenario-Level Distribution Shift}

\author{San Kim}
\orcid{0009-0005-1636-775X}
\affiliation{%
  \institution{Sungkyunkwan University}
  \city{Suwon}
  \country{Republic of Korea}}  
\email{kimsan1120@g.skku.edu}

\author{JinYeong Bak}
\orcid{0000-0002-3212-5241}
\affiliation{%
  \institution{Sungkyunkwan University}
  \city{Suwon}
  \country{Republic of Korea}}  
\email{jy.bak@skku.edu}

\renewcommand{\shortauthors}{Kim and Bak}

\begin{abstract}
Conventional in-distribution evaluation can overestimate robustness when training and test data share recurring task-specific patterns or surface cues. This risk is especially relevant in social-engineering fraud detection, where attackers can preserve malicious intent while changing the scenario, impersonated entity, or wording. We study this problem as scenario-level out-of-distribution (SL-OOD) detection for SMS and voice phishing, where entire attack scenarios are held out from training while the label space remains fixed. This setting tests whether models can generalize to unseen attack scenarios using decision-relevant evidence rather than familiar scenario-specific cues.
Using this SL-OOD evaluation, we find that high in-distribution performance does not reliably predict held-out robustness across feature-, encoder-, and decoder-based baselines. We interpret this gap as \emph{scenario memorization}: reliance on recurring scenario-specific lexical or entity cues rather than decision-relevant evidence. We propose ECoG, an evidence-consistent generative framework that combines evidence-span supervision with a rationale--label consistency objective during training. On the 0.5B decoder, relative to the same backbone trained without consistency regularization, ECoG raises Macro-F1 on OOD challenging instances by \(3.22\) points, reduces the share of predictions whose generated rationale supports the opposite label by \(4.22\) points, and increases token-level overlap with reference evidence spans by \(8.38\) points; the reduction in prediction--rationale inconsistency is consistent across four decoder backbones. These results suggest that compact generative detectors can benefit from evidence supervision and rationale--label consistency under social-engineering shift.\footnote{Code and Data: \url{https://github.com/kimsan1120/ECoG}}
\begingroup\def\thefootnote{$\dagger$}\footnotetext{JinYeong Bak is a corresponding author}\endgroup
\renewcommand{\thefootnote}{\arabic{footnote}}
\end{abstract}



\maketitle

\section{Introduction}
Conventional in-distribution evaluation can overestimate robustness when training and test data share recurring task-specific patterns or surface cues~\cite{Geirhos_2020, mccoy-etal-2019-right, hendrycks-etal-2020-pretrained}. 
This risk is acute in social-engineering fraud, where attackers can preserve malicious intent while changing the scenario, impersonated entity, or wording~\cite{cho-seo-2025-towards}.
In SMS and voice phishing, a detector trained and tested on recurring scenarios may therefore achieve high accuracy by recognizing familiar templates rather than decision-relevant evidence.

We frame SMS and voice phishing detection as scenario-level out-of-distribution (SL-OOD) detection. Entire social-engineering scenarios are held out from training while the label space remains fixed. Evaluation thus asks whether detectors generalize to unseen attack scenarios rather than familiar scenario-specific cues. This setting is stricter than a conventional random split because train and test examples no longer share the same scenario-specific signals. We refer to the associated failure mode as \emph{scenario memorization}: reliance on recurring scenario-specific lexical or entity cues rather than decision-relevant evidence, used descriptively rather than as a claim about internal mechanisms.

For generative detectors, scenario-level robustness is not only a label-prediction problem. Under held-out scenarios, a model may produce a confident label with a fluent rationale that is weakly supported by the input or directionally inconsistent with the prediction. We therefore pair SL-OOD evaluation with a challenging set designed to stress intent-level discrimination, and evaluate both classification robustness and generated-output behavior: evidence-span overlap and prediction--rationale consistency.

We propose ECoG, an Evidence-Consistent Generative Framework for decoder-based detection under unseen social-engineering scenarios. ECoG combines evidence-span supervision with rationale--label consistency regularization during training, encouraging input-side grounding and label-consistent rationales without changing the inference-time decoding path. Unlike prior data-curation or inference-time approaches, ECoG applies consistency as an auxiliary loss within single-stage fine-tuning.

Experiments across feature-, encoder-, and decoder-based baselines confirm that in-distribution performance is a weak predictor of scenario-level robustness, while ECoG improves hard-case classification and generated-output behavior across decoder backbones.

We make three contributions. \textbf{First}, we introduce an SL-OOD holdout protocol and challenging set for evaluating detectors beyond conventional in-distribution splits. \textbf{Second}, we propose ECoG, a decoder-based generative detector that combines evidence-span supervision with a training-time rationale--label consistency regularizer. \textbf{Third}, we show that conventional model selection can be misleading under unseen scenarios, while evidence supervision with rationale--label consistency consistently reduces prediction--rationale inconsistency across decoder backbones and improves hard-case classification and reference evidence-span overlap in a backbone-dependent manner.
\section{Related Work}
\subsection{Distribution Shift and Shortcut Learning}

Models trained and evaluated under the independent and identically distributed (i.i.d.) assumption can achieve high test performance by relying on dataset-specific regularities rather than task-relevant evidence. This behavior is commonly described as shortcut learning: decision rules that work well under standard evaluation but fail to transfer to more challenging or shifted conditions~\cite{Geirhos_2020}. In NLP, such failures have been observed when models rely on shallow lexical or syntactic heuristics~\cite{mccoy-etal-2019-right}. These concerns motivate evaluation beyond random splits, including contrast sets~\cite{gardner-etal-2020-evaluating}, behavioral testing~\cite{ribeiro-etal-2020-beyond}, and counterfactually augmented data~\cite{Kaushik2020Learning}.

Prior work characterizes generalization beyond i.i.d. test sets through taxonomies of generalization and distribution shift~\cite{hupkes-etal-2023-taxonomy,arora-etal-2021-types}. Under this view, scenario-level holdout can be viewed as a form of covariate shift: the input distribution changes while the labeling function remains fixed. Related work also shows that pretrained Transformers improve, but do not fully solve, OOD robustness~\cite{hendrycks-etal-2020-pretrained}.

A close benchmark-level analog is misinfo-general~\cite{verhoeven-et-al-2025-yesterday}, which evaluates classifiers along explicit shift axes in a rapidly changing text domain. Our SL-OOD protocol follows this motivation by holding out entire social-engineering scenarios rather than random examples. In this setting, the corresponding shortcut appears as \emph{scenario memorization}: reliance on recurring scenarios, impersonated entities, or templates rather than decision-relevant evidence.

\subsection{Evidence and Rationale Supervision}
Extractive rationalization is often formulated as selecting a sparse subset of the input that supports the predicted label~\cite{lei-etal-2016-rationalizing}. ERASER~\cite{deyoung-etal-2020-eraser} consolidated this line with human rationale annotations and rationale evaluation metrics, while Jacovi and Goldberg~\cite{jacovi-goldberg-2020-towards} distinguish plausible human-readable rationales from explanations that are faithful to the model's actual decision process.

Beyond extractive rationales, free-text explanations have been used as complementary supervision signals. e-SNLI~\cite{NEURIPS2018_4c7a167b} studies models trained to generate natural-language explanations, and subsequent work examines whether rationalization or human explanations can improve robustness~\cite{chen-etal-2022-rationalization, Stacey_Belinkov_Rei_2022}. Evidence-based verification similarly uses evidence and justifications to support classification decisions~\cite{atanasova-etal-2020-generating-fact,thorne-etal-2018-fact}. We build on this line by using evidence spans and rationales as supervision signals, while evaluating generated evidence through reference-span overlap rather than claiming causal faithfulness.

\subsection{Rationale--Label Consistency}

Decoder-based language models can generate labels and natural-language rationales along a single decoding path, as in text-to-text self-rationalization approaches such as WT5~\cite{Narang2020WT5TT}. However, joint generation alone does not ensure faithful or label-consistent rationales; generated reasoning can misrepresent the basis of model predictions~\cite{turpin2023language}.

Prior work studies this gap through label-specific explanations, label--rationale association, simulatability, and counterfactual consistency. NILE~\cite{kumar-talukdar-2020-nile} makes predictions through label-specific explanations, while Wiegreffe et al.~\cite{wiegreffe-etal-2021-measuring} test label--rationale association. Hase et al.~\cite{hase-etal-2020-leakage} and REV~\cite{chen-etal-2023-rev} evaluate whether rationales provide useful label-relevant information, and PINTO~\cite{wang2022pinto} and SCOTT~\cite{wang-etal-2023-scott} use counterfactual objectives to encourage models to respect rationales. Most related to our objective, Atanasova et al.~\cite{atanasova-etal-2023-faithfulness} propose input reconstruction as a post-hoc diagnostic, and Veerubhotla et al.~\cite{veerubhotla-etal-2023-shot} encourage explanations to depend on predicted labels through masked label regularization. Recent work also uses consistency or reward signals to improve rationale quality: CREST evaluates generated rationales through consistency-driven follow-up questions for self-training~\cite{lee-etal-2025-self}, while Drift uses dual-reward probabilistic inference to improve rationale faithfulness~\cite{li-etal-2025-drift}. ECoG differs from these data-curation and inference-time approaches by applying rationale--label consistency as a parameter-level auxiliary loss during single-stage fine-tuning, while leaving inference-time decoding unchanged.

\subsection{Textual Phishing Detection}
Prior work on SMS and voice phishing detection spans feature-based classifiers, neural sequence models, pretrained language models, and recent large language model (LLM)-based approaches. Early and practical systems rely on surface-form features such as bag-of-words, TF--IDF, character \(n\)-grams, and handcrafted lexical or entity-level cues, often combined with lightweight machine-learning models~\cite{realtime}. In Korean voice phishing detection, attention-based CNN--BiLSTM models and NER/key-tag pipelines have been used to capture discriminative lexical and entity patterns~\cite{math11143217,10496052}. More recent systems incorporate multimodal audio--text signals or data augmentation to improve practical coverage~\cite{app152011170,10901962,11142247}. These studies establish the practical importance of phishing detection across modalities, but much of this line evaluates models under task-specific or in-distribution conditions where recurring scenario cues may be shared across train and test splits.

With pretrained language models, phishing detection has also been reframed as supervised sequence classification or generative classification. KorSmishing Explainer~\cite{lee-han-2024-korsmishing} pairs Korean smishing detection with natural-language explanation generation, while recent practical phishing detection work introduces scenario-partitioned evaluation and adaptation methods such as PEFT and TAPT for OOD robustness~\cite{cho-seo-2025-towards}. We extend this application line by evaluating Korean SMS messages and ASR-transcribed voice-phishing utterances under SL-OOD splits and by coupling detection with evidence and rationale supervision.

\paragraph{Positioning of our work.}
Prior work separately studies shortcut-based generalization failures, evidence and rationale supervision, rationale--label consistency, and phishing detection. ECoG combines these threads in an SL-OOD detection framework, using evidence supervision and rationale--label consistency regularization to encourage input-side grounding and prediction--rationale consistency under held-out scenarios.
\section{Dataset and Evaluation Protocol}
\subsection{Dataset Construction}
\label{sec:data:construction}
\setlength{\textfloatsep}{8pt plus 1pt minus 2pt}

\begin{table}[t!]
\centering
\small
\renewcommand{\arraystretch}{0.8}
\setlength{\tabcolsep}{6pt}

\newcommand{\schead}[1]{\hspace*{1pt}#1}
\newcommand{\scen}[1]{\hspace*{4pt}#1}

\caption{Dataset composition by split, modality, and scenario. Finance appears in both SMS and Voice.}

\label{tab:dataset_stats}

\begin{tabular}{
@{}
>{\centering\arraybackslash}p{0.15\columnwidth}
>{\centering\arraybackslash}p{0.15\columnwidth}
>{\raggedright\arraybackslash}p{0.19\columnwidth}
>{\raggedleft\arraybackslash}p{0.080\columnwidth}
>{\raggedleft\arraybackslash}p{0.090\columnwidth}
>{\raggedleft\arraybackslash}p{0.095\columnwidth}
@{}
}
\toprule
\textbf{Split} 
& \textbf{Modality} 
& \schead{\textbf{Scenario}}
& \textbf{Phish.}
& \textbf{Benign}
& \textbf{Total} \\
\midrule

\multirow{6}{*}{\textbf{Train}}
& \multirow{3}{*}{SMS}   & \scen{Credit}     & 8,153 & 7,462 & 15,615 \\
&                        & \scen{Finance}    & 3,657 & 8,302 & 11,959 \\
&                        & \scen{Parcel}     & 7,147 & 5,309 & 12,456 \\
\cmidrule(l){2-6}
& \multirow{2}{*}{Voice} & \scen{Government} &   888 &   818 &  1,706 \\
&                        & \scen{Finance}    & 2,233 & 1,720 &  3,953 \\
\midrule

\multirow{6}{*}{\textbf{Validation}}
& \multirow{3}{*}{SMS}   & \scen{Credit}     & 1,559 & 1,006 &  2,565 \\
&                        & \scen{Finance}    &   466 & 1,111 &  1,577 \\
&                        & \scen{Parcel}     & 1,243 &   716 &  1,959 \\
\cmidrule(l){2-6}
& \multirow{2}{*}{Voice} & \scen{Government} &   114 &   104 &    218 \\
&                        & \scen{Finance}    &   282 &   216 &    498 \\
\midrule

\multirow{6}{*}{\textbf{Test}}
& \multirow{3}{*}{SMS}   & \scen{Credit}     & 1,572 & 1,014 &  2,586 \\
&                        & \scen{Finance}    &   467 & 1,113 &  1,580 \\
&                        & \scen{Parcel}     & 1,243 &   707 &  1,950 \\
\cmidrule(l){2-6}
& \multirow{2}{*}{Voice} & \scen{Government} &   113 &   102 &    215 \\
&                        & \scen{Finance}    &   283 &   214 &    497 \\
\midrule

\multirow{6}{*}{\textbf{Challenging}}
& \multirow{3}{*}{SMS}   & \scen{Credit}     &   112 &    91 &    203 \\
&                        & \scen{Finance}    &   101 &   100 &    201 \\
&                        & \scen{Parcel}     &    98 &    53 &    151 \\
\cmidrule(l){2-6}
& \multirow{2}{*}{Voice} & \scen{Government} &   447 &   459 &    906 \\
&                        & \scen{Finance}    &   224 &   231 &    455 \\
\bottomrule
\end{tabular}
\end{table}

\paragraph{Data sources and reconstruction.}
We construct a Korean phishing detection dataset with two modalities: SMS messages and voice-call transcripts. The raw collection builds on prior work~\cite{cho-seo-2025-towards}. Benign samples are sourced from DeepNatural\footnote{\url{https://startups.koraia.org/company/81}} and AIHub.\footnote{\url{https://www.aihub.or.kr}} We do not treat this collection as an off-the-shelf benchmark: we reconstruct it for scenario-level OOD evaluation through deduplication, scope filtering, scenario annotation, and challenging-set construction. More detailed criteria are documented in the supplementary repository (Appendix~\ref{app:repro}).

\paragraph{Samples and scenarios.}
Phishing samples originate from institutionally verified Korean phishing cases associated with the prior collection~\cite{cho-seo-2025-towards}. Benign SMS messages are drawn from DeepNatural, and benign voice transcripts are drawn from AIHub civil-complaint call-center data. To reduce topic-level shortcuts, we select benign samples that mention entities commonly impersonated in phishing scenarios, making benign examples lexically closer to phishing messages. Each instance consists of an input text, a binary phishing label, and a scenario label. The dataset contains four scenarios: \textbf{Finance}, \textbf{Parcel}, \textbf{Credit}, and \textbf{Government}; Table~\ref{tab:dataset_stats} reports the sample counts.

\paragraph{Evidence and rationale annotation.}
We use \textit{GPT-4o-mini}~\cite{openaigpt4omini} with temperature \(0\) to construct evidence-span and rationale annotations for training supervision. Evidence spans are required to be copied verbatim from the input. Annotation filtering, quality audit, PII de-identification, prompts, and reproducibility details are provided in Appendix~\ref{appendix:annotation}.

\subsection{Evaluation Protocols}
\label{sec:evaluation_protocols}

\paragraph{In-distribution evaluation.}
The in-distribution setting follows a conventional random split, where training, validation, and test sets share the same scenario distribution.

\paragraph{Scenario-level holdout evaluation.}
In the scenario-level holdout setting, one or more phishing scenarios are excluded from training and used only for evaluation. This tests whether detectors recognize phishing intent under social-engineering scenarios not observed during training. For SMS, we hold out \textbf{Credit}, \textbf{Finance}, and \textbf{Parcel} in turn; for Voice, we hold out \textbf{Government} and \textbf{Finance} in turn. OOD scores in Table~\ref{tab:headline} are scenario-balanced averages across folds within each modality.

\paragraph{Challenging evaluation set.}
We construct a challenging set of ambiguous or lexically confusable instances that require intent-level discrimination. It is used as a stress test rather than an adversarial attack benchmark. SMS candidates are mined with encoder baselines but retained only after blind human difficulty filtering; construction details and retention criteria are provided in Appendix~\ref{appendix:challenging_set}.

\begin{figure*}[t!]
  \centering
  \includegraphics[width=\textwidth]{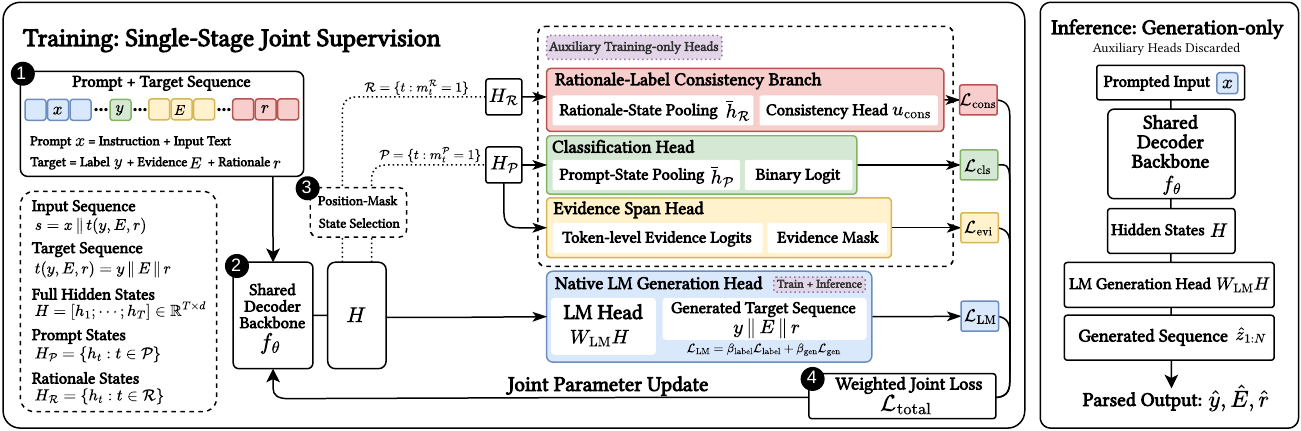}
\caption{Overview of ECoG.
(1) The training sequence consists of the prompted input \(x\) and target sequence \(y \Vert E \Vert r\).
(2) A shared decoder produces hidden states \(H\).
(3) Position masks select prompt and rationale states for auxiliary training-only heads: classification, evidence-span supervision, and rationale--label consistency.
(4) The native LM head supervises the generated target sequence through \(\mathcal{L}_{\mathrm{LM}}=\beta_{\mathrm{label}}\mathcal{L}_{\mathrm{label}}+\beta_{\mathrm{gen}}\mathcal{L}_{\mathrm{gen}}\), and all objectives are combined in the weighted joint loss.
At inference, all auxiliary heads are discarded, and the decoder generates the label, evidence spans, and rationale from \(x\) using only the native LM head.}
\Description{Two-panel schematic of the ECoG framework. The left panel, labeled Training: Single-Stage Joint Supervision, shows four numbered steps. Step 1 is a prompt and target sequence, where the prompt is an instruction plus input text and the target is a label, evidence spans, and a rationale. Step 2 is a shared decoder backbone that produces hidden states H. Step 3 is a position-mask state selection block that routes prompt states and rationale states to three auxiliary training-only heads, enclosed in a dashed box: a rationale-label consistency branch, a classification head, and an evidence span head, producing the consistency, classification, and evidence losses. A separate native LM generation head, used at both training and inference, produces the LM loss from the generated target sequence. Step 4 combines all terms into a weighted joint loss that drives a joint parameter update of the shared backbone. The right panel, labeled Inference: Generation-only, shows the auxiliary heads discarded: the prompted input passes through the shared decoder backbone and the LM generation head to produce a generated sequence, which is parsed into a predicted label, evidence spans, and rationale.}
  \label{fig:ecog-overview}
\end{figure*}

\section{The ECoG Framework}
\label{sec:method}
We describe ECoG, a decoder-based detector that jointly generates a label, evidence spans, and a rationale with evidence supervision and rationale--label consistency.

\subsection{Problem Formulation}
\label{sec:method:problem}

Let \(x\) denote the prompted input sequence, including the task instruction and the message or transcript text, and let \(y \in \{0,1\}\) denote the binary label. For supervision, each instance is paired with a set of reference evidence spans \(E=\{e_1,\dots,e_K\}\), where each \(e_k\) is a contiguous substring of the message or transcript text. Phishing instances contain at least one evidence span, while benign instances may have an empty evidence set when no explicit benign-supporting cue is present. Each instance is also paired with a natural-language rationale \(r\). During training, the model observes \((x,y,E,r)\) and learns a decoder-based generative model \(f_\theta\). At inference time, \(f_\theta\) receives only \(x\) and generates a structured output consisting of a label \(\hat{y}\), evidence-span surface forms \(\hat{E}\), and a natural-language rationale \(\hat{r}\).

\subsection{Framework Overview}
\label{sec:method:overview}

Scenario-level distribution shift motivates three design goals: predictions should rely less on lexical or entity cues from seen scenarios, decisions should use input-side evidence, and rationales should remain directionally consistent with the predicted label.
Figure~\ref{fig:ecog-overview} summarizes the training and inference flow of ECoG.

During training, the decoder backbone \(f_\theta\) processes the prompt--target sequence \(x \Vert t(y,E,r)\), where \(\Vert\) denotes string concatenation and the target serialization \(t(\cdot)\) is defined in Section~\ref{sec:method:io}, producing hidden states \(H \in \mathbb{R}^{T \times d}\) with sequence length \(T\) and hidden size \(d\).

Position masks select prompt states \(H_{\mathcal{P}}\) for classification and evidence supervision, and rationale states \(H_{\mathcal{R}}\) for the consistency branch, where \(\mathcal{P}\) and \(\mathcal{R}\) denote the prompt and rationale token positions defined in Section~\ref{sec:method:io}. The objective terms are defined below and combined in Section~\ref{sec:method:training}.

\subsection{Input and Target Construction}
\label{sec:method:io}

\paragraph{Prompted input.}
We construct \(x\) by wrapping the message or transcript text with a fixed instruction that directs the model to predict the label, decision-supporting evidence spans, and a Korean rationale in a structured target format. For reproducibility, we provide both the original Korean prompt and its English translation in the supplementary repository. We denote the set of token positions occupied by the prompted input \(x\) as \(\mathcal{P} \subset \{1, \dots, T\}\). Thus, the full training sequence is \(s = x \Vert t(y,E,r)\).

\paragraph{Target.}
The target \(t(y,E,r)\) is a single string composed of three contiguous fields: a label field, an evidence field, and a rationale field. The label field is placed first, followed by evidence and rationale fields separated by fixed delimiters:
\begin{equation}
t(y,E,r)
=
\underbrace{y}_{\text{label}}
\;\Vert\;
\underbrace{\delta_{\mathrm{evi}} \;\Vert\; \mathrm{seq}(E)}_{\text{evidence}}
\;\Vert\;
\underbrace{\delta_{\mathrm{exp}} \;\Vert\; r}_{\text{rationale}},
\end{equation}
where \(\delta_{\mathrm{evi}}\) is the evidence delimiter, and \(\delta_{\mathrm{exp}}\) is the rationale delimiter. The function \(\mathrm{seq}(E)\) serializes the evidence spans as a list; if \(E\) is empty, it is replaced by a fixed ``none'' marker.

We write the resulting target-token positions as \(\mathcal{G} \subset \{1,\dots,T\}\) and let \(t_\ell \in \mathcal{G}\) denote the label-token position. Since the label field is placed first, \(t_\ell\) is the first generated target position.\footnote{We verified that the two label verbalizers used in our experiments are each represented as a single token in all backbone tokenizers.} We further define \(\mathcal{R} \subset \mathcal{G}\) as the set of token positions inside the rationale field, excluding the rationale delimiter. Placing the label field first provides a direct supervision signal for the binary decision at the earliest generated position; we examine this design choice empirically in Section~\ref{sec:exp:ablation}.

\paragraph{Evidence token mask.}
To supervise the evidence span head, we project the span-level annotation \(E\) onto the prompt token positions by setting \(m^{\mathrm{evi}}_t=1\) for each \(t\in\mathcal{P}\) whose subword character span overlaps any \(e_k\in E\), and \(0\) otherwise. We also define a loss mask \(m^{\mathrm{loss}}_t\) that keeps all evidence tokens and a sampled subset of non-evidence tokens, reducing the imbalance between evidence and non-evidence positions.

\subsection{Generation Objectives}
\label{sec:method:gen}

Both generation objectives use the native LM head \(W_{\mathrm{LM}} \in \mathbb{R}^{|V| \times d}\) applied to the decoder hidden states \(H\), where \(V\) is the model vocabulary. We denote the LM logits at position \(t\) by \(\ell_t = W_{\mathrm{LM}}h_t \in \mathbb{R}^{|V|}\).

\paragraph{Label-token Prediction.}
At position \(t_\ell-1\), whose output predicts the label token at \(t_\ell\), we restrict the softmax to two label verbalizer tokens \(\{v_0,v_1\} \subset V\) and apply cross-entropy against the label:
\begin{equation}
\mathcal{L}_{\mathrm{label}}
=
\mathrm{CE}\!\left(
\ell_{t_\ell-1}\big|_{\{v_0,v_1\}},
y
\right).
\end{equation}
This term supervises label-token prediction on the native LM generation path.

\paragraph{Evidence and rationale generation.}
For the remaining target positions \(\mathcal{G}'=\mathcal{G}\setminus\{t_\ell\}\), we apply standard next-token cross-entropy:
\begin{equation}
\mathcal{L}_{\mathrm{gen}}
=
-\frac{1}{|\mathcal{G}'|}
\sum_{t\in\mathcal{G}'}
\log p_\theta(z_t \mid x, z_{<t}),
\end{equation}
where \(z_t\) denotes the target token at position \(t\) and
\(p_\theta(\cdot \mid x,z_{<t})=\mathrm{softmax}(\ell_{t-1})\). This term supervises the serialized evidence and rationale fields.

We group the two native LM-head losses as
\begin{equation}
\mathcal{L}_{\mathrm{LM}}
=
\beta_{\mathrm{label}}\mathcal{L}_{\mathrm{label}}
+
\beta_{\mathrm{gen}}\mathcal{L}_{\mathrm{gen}} .
\end{equation}
Here, \(\mathcal{L}_{\mathrm{label}}\) supervises label-token prediction, while \(\mathcal{L}_{\mathrm{gen}}\) supervises the remaining generated target tokens.

\subsection{Auxiliary Training-Only Supervision}
\label{sec:method:aux}

The classification and evidence span heads are auxiliary training-only heads computed from \(H_{\mathcal{P}} = \{h_t : t \in \mathcal{P}\}\).

\paragraph{Auxiliary prompt-state classification.}
We add a training-only auxiliary classifier over prompt-position hidden states, complementing label-token prediction through the native LM head. We mean-pool these states as
\(\bar{h}_{\mathcal{P}} = |\mathcal{P}|^{-1}\sum_{t\in\mathcal{P}} h_t\),
compute a scalar logit
\(u_{\mathrm{cls}} = W_{\mathrm{cls}}\bar{h}_{\mathcal{P}} + b_{\mathrm{cls}}\),
and apply weighted binary cross-entropy:
\[
\mathcal{L}_{\mathrm{cls}}
=
\mathrm{BCEWithLogits}(u_{\mathrm{cls}}, y;\, w_+),
\]
where \(w_+\) is the positive-class weight for the corresponding training split.

\paragraph{Evidence span head.}
For each prompt token \(t \in \mathcal{P}\), the evidence head produces a scalar logit \(s^{\mathrm{evi}}_t = W_{\mathrm{evi}} h_t + b_{\mathrm{evi}}\), with \(W_{\mathrm{evi}} \in \mathbb{R}^{1 \times d}\). The loss is a masked token-level binary cross-entropy:
\begin{equation}
\mathcal{L}_{\mathrm{evi}}
=
\frac{1}{\sum_t m^{\mathrm{loss}}_t}
\sum_{t \in \mathcal{P}}
m^{\mathrm{loss}}_t \cdot
\mathrm{BCEWithLogits}\!\left(s^{\mathrm{evi}}_t,\; m^{\mathrm{evi}}_t\right).
\end{equation}
When the evidence field is ``none'', we omit \(\mathcal{L}_{\mathrm{evi}}\) for that instance. The LM head is still trained to generate the serialized evidence field, while the evidence span head provides input-side localization supervision for instances with annotated spans.

\subsection{Rationale--Label Consistency Learning}
\label{sec:method:cons}

The consistency branch adds a training-time regularizer by predicting the label from rationale-position hidden states.

\paragraph{Rationale-position hidden states.}
We apply the consistency objective only to rationale-token hidden states
\(H_{\mathcal{R}}=\{h_t:t\in\mathcal{R}\}\). The pooling mask excludes the label-token position, evidence delimiters, and evidence-span tokens, so the consistency head does not directly pool hidden states from the label or evidence fields. However, because \(f_\theta\) is a causal decoder trained under teacher forcing, rationale-position states can still be contextually conditioned on preceding target tokens. We therefore treat this branch as a training-time regularizer over decoder states, not as evidence that the generated rationale is causally faithful to the prediction.

\paragraph{Consistency head.}
We mean-pool the selected rationale states,
\(\bar{h}_{\mathcal{R}} = |\mathcal{R}|^{-1}\sum_{t\in\mathcal{R}} h_t\),
and apply a two-way linear head,
\(u_{\mathrm{cons}} = W_{\mathrm{cons}}\bar{h}_{\mathcal{R}} + b_{\mathrm{cons}} \in \mathbb{R}^{2}\),
where \(W_{\mathrm{cons}}\in\mathbb{R}^{2\times d}\). The consistency loss is
\[
\mathcal{L}_{\mathrm{cons}}=\mathrm{CE}(u_{\mathrm{cons}},y).
\]
Its gradients update the shared decoder states selected by the rationale-position mask, while the head parameters remain separate from the LM head.

\subsection{Joint Training Objective and Inference}
\label{sec:method:training}

\paragraph{Joint objective.}
We combine the native LM loss with the auxiliary training-only objectives:
\begin{equation}
\mathcal{L}_{\mathrm{total}}
=
\mathcal{L}_{\mathrm{LM}}
+
\beta_{\mathrm{cls}}\mathcal{L}_{\mathrm{cls}}
+
\beta_{\mathrm{evi}}\mathcal{L}_{\mathrm{evi}}
+
\lambda\mathcal{L}_{\mathrm{cons}} .
\end{equation}
Here, \(\mathcal{L}_{\mathrm{LM}}\) denotes the weighted native LM-head supervision, while \(\mathcal{L}_{\mathrm{cls}}\), \(\mathcal{L}_{\mathrm{evi}}\), and \(\mathcal{L}_{\mathrm{cons}}\) are auxiliary training-only objectives.
We keep the base loss weights \(\beta_{\mathrm{label}}, \beta_{\mathrm{gen}}, \beta_{\mathrm{cls}}, \beta_{\mathrm{evi}}\) fixed across all experiments; their values and other training hyperparameters are provided in Appendix~\ref{app:config}.
We sweep \(\lambda \in \{0.0, 0.05, 0.1, 0.2\}\) as a sensitivity analysis, with \(\lambda=0\) corresponding to Joint w/o C (Section~\ref{sec:exp:ablation}).
We fix \(\lambda=0.1\) as a single default across backbones and scenarios, rather than selecting \(\lambda\) on held-out OOD performance.

\paragraph{Training procedure.}
We train ECoG with single-stage full fine-tuning, jointly updating the shared decoder backbone, the native LM head, and the auxiliary training-only heads. Each step performs one forward pass, computes all loss terms from the shared hidden states, and backpropagates \(\mathcal{L}_{\mathrm{total}}\). The auxiliary heads add fewer than \(0.1\%\) of the backbone parameter count. Other training details are summarized in Appendix~\ref{app:config}.

\paragraph{Inference.}
At inference time, ECoG decodes from the prompted input \(x\) alone using only the decoder backbone and native LM head. The auxiliary heads are discarded entirely and are never queried for label prediction, evidence extraction, or rationale generation. The first decoding step is restricted to the two label verbalizers, and the higher-scoring token determines \(\hat{y}\). The remaining output is parsed into \(\hat{E}\) and \(\hat{r}\) using the fixed delimiters \(\delta_{\mathrm{evi}}\) and \(\delta_{\mathrm{exp}}\). These fields are used for evidence-overlap and prediction--rationale consistency analyses in Section~\ref{sec:exp:explanation}.

\section{Experiments}
\label{sec:experiments}

We structure the experiments around four research questions, stated as the headings of Sections~\ref{sec:exp:rq1}--\ref{sec:exp:explanation}. Table~\ref{tab:headline} and Figure~\ref{fig:id_ood_gap} address RQ1--RQ2; Table~\ref{tab:scale} reports larger-backbone results for RQ2; Table~\ref{tab:mechanism} and Figure~\ref{fig:lambda_sweep} address RQ3--RQ4.

\subsection{Experimental Setup} 
\label{sec:exp:setup}

\paragraph{Baselines.}
We compare ECoG against six baseline groups covering distinct modeling assumptions:
(i) a feature-based classifier (TF-IDF + SVM)~\cite{realtime};
(ii) task-trained neural classifiers (BiLSTM~\cite{10901962} and CNN--BiLSTM with attention~\cite{math11143217});
(iii) an encoder PLM classifier (KoBERT with TAPT and LoRA adaptation)~\cite{cho-seo-2025-towards};
(iv) proprietary LLM baselines (GPT-5.4~\cite{openaigpt54} and Gemini 3.1 Flash-Lite~\cite{googlegemini31}), evaluated with the same task instruction but without task-specific fine-tuning or in-context examples;\footnote{For proprietary LLMs, ID and OOD denote the evaluation split rather than the training regime; these models are not fine-tuned on our ID training split.}
(v) a decoder LLM trained with QLoRA (KULLM-5B)~\cite{lee-han-2024-korsmishing}; and
(vi) decoder SLM baselines using the same backbone families as ECoG but simpler objectives, including single-token supervised fine-tuning (ST-SFT), rationale-only generation.
All baselines are evaluated under the same scenario-level holdout protocol (Section~\ref{sec:evaluation_protocols}).

\begin{table*}[t!]
\centering
\caption{Macro-F1, scenario-balanced within each modality; bold and underline mark the best and second-best in each column. Voice Test ID saturation (e.g., HCX-0.5B at \(100.00\)) reflects the limited size of the random-split Voice test set; OOD and Challenging columns are more discriminative.}

\label{tab:headline}
\small
\setlength{\tabcolsep}{2.5pt}
\renewcommand{\arraystretch}{0.9}
\centering
\resizebox{0.98\textwidth}{!}{%
\begin{tabular}{
>{\centering\arraybackslash}m{0.125\textwidth}
>{\raggedright\arraybackslash}p{0.185\textwidth}
*{8}{>{\raggedleft\arraybackslash}p{0.052\textwidth}}
}
\toprule

\multirow{4}{*}{\makecell[c]{\textbf{Type}}}
& \multirow{4}{*}{\makecell[c]{\textbf{Model / Variant}}}
& \multicolumn{4}{c}{\textbf{SMS}}
& \multicolumn{4}{c}{\textbf{Voice}} \\
\cmidrule(lr){3-6} \cmidrule(lr){7-10}
& & \multicolumn{2}{c}{Test}
& \multicolumn{2}{c}{Challenging}
& \multicolumn{2}{c}{Test}
& \multicolumn{2}{c}{Challenging} \\
\cmidrule(lr){3-4} \cmidrule(lr){5-6}
\cmidrule(lr){7-8} \cmidrule(lr){9-10}
& & ID & OOD & ID & OOD & ID & OOD & ID & OOD \\
\midrule

Feature-based
& TF-IDF + SVM~\cite{realtime}
& 97.29 & 74.92 & 66.92 & 40.13 & 99.13 & 80.33 & 88.15 & 65.95 \\

\addlinespace[1pt]
\multirow{2}{*}{Neural}
& BiLSTM~\cite{10901962}
& 98.36 & 86.07 & 80.17 & 58.19 & 98.15 & 85.68 & 88.97 & 74.81 \\
& CNN-BiLSTM + Attn.~\cite{math11143217}
& 98.60 & 91.81 & 81.48 & 70.32 & 99.13 & 86.40 & 90.11 & 77.72 \\

\addlinespace[1pt]
Encoder PLM
& KoBERT~\cite{cho-seo-2025-towards}
& 95.70 & 89.15 & 74.96 & 72.42 & 77.77 & 45.57 & 60.64 & 44.46 \\

Decoder LLM
& KULLM-5B~\cite{lee-han-2024-korsmishing}
& 97.82 & 92.39 & 95.55 & 86.88 & 99.25 & 90.95 & 89.75 & 76.08 \\

\midrule
\multirow{4}{*}{\makecell[c]{Decoder SLM\\Baseline}}
& HCX-0.5B ST-SFT
& 98.52 & 91.00 & 92.09 & 84.98 & 99.10 & 91.62 & 93.84 & 81.62 \\
& Qwen3-0.6B ST-SFT
& 98.02 & 90.87 & 89.24 & 81.01 & 98.57 & 85.44 & 91.71 & 71.25 \\
& HCX-0.5B Rat.-only
& 98.48 & 91.22 & 92.86 & 80.26 & 99.25 & 87.27 & 92.76 & 76.42 \\
& Qwen3-0.6B Rat.-only
& 98.17 & 93.30 & 87.62 & 76.07 & 97.71 & 62.84 & 86.08 & 59.90 \\

\midrule
\multirow{2}{*}{\makecell[c]{Proprietary LLM}}
& Gemini 3.1 Flash-Lite
& 83.19 & 74.50 & 39.97 & 42.02 & 93.73 & 85.84 & 92.32 & 82.11 \\
& GPT-5.4
& 81.43 & 80.67 & 41.74 & 33.74 & 82.55 & 82.42 & 90.60 & \textbf{90.42} \\

\midrule
\multirow{2}{*}{\textbf{ECoG (ours)}}
& \textbf{HCX-0.5B}
& \textbf{99.65} & \textbf{94.97} & \underline{96.80} & \textbf{92.71} & \textbf{100.00} & \textbf{94.56} & \textbf{97.36} & \underline{88.98} \\
& \textbf{Qwen3-0.6B}
& \underline{99.40} & \underline{94.93} & \textbf{97.88} & \underline{89.75} & \underline{99.66} & \underline{92.05} & \underline{95.66} & 84.78 \\

\bottomrule
\end{tabular}
}
\end{table*}

\paragraph{Backbones.}
ECoG uses two Korean-capable decoder backbone families: HyperCLOVA X SEED (0.5B and 1.5B) and Qwen3 (0.6B and 1.7B). We focus on compact SLM-scale backbones to study whether resource-efficient decoders can provide robust detection and explanation behavior under scenario-level shift, rather than relying only on large proprietary LLMs. We use the corresponding instruction-tuned checkpoints and tokenizers for each model size.

\paragraph{Training and checkpointing.}
All task-trained models are trained for a fixed number of epochs and reported using the final checkpoint. Validation loss is monitored only for diagnostics: we do not use early stopping, validation-based checkpoint selection, or held-out scenario results for model selection. Held-out test scenarios and challenging sets are used only for final evaluation. Training hyperparameters and checkpoint details are provided in Appendix~\ref{app:config}.

\paragraph{Metrics.}
We use Macro-F1 as the main classification metric for the split-level results in Table~\ref{tab:headline}. For ablation analyses, we summarize performance with two aggregate metrics defined below: condition-balanced Macro-F1 (\textbf{Cls F1}) and OOD-challenging Macro-F1 (\textbf{Hard F1}). To evaluate the generated outputs, we report token-level span F1 against reference evidence spans (\textbf{Span F1}), BERTScore F1 against reference rationales (\textbf{BERTScore}), and prediction--rationale inconsistency rate (\textbf{Inc.}).

\paragraph{Prediction--rationale inconsistency (Inc.).}
For each evaluation instance, we compare the generated label \(\hat{y}\) with the generated rationale \(\hat{r}\). We use a rule-based cue parser to assign each rationale a direction: \emph{benign-only}, \emph{phishing-only}, \emph{mixed}, or \emph{none}. The parser detects lexical cues associated with each rationale direction, with simple negation handling to avoid assigning negated cues to the wrong direction. An instance is counted as inconsistent when the generated label and rationale direction disagree: a phishing label with a benign-only rationale, or a benign label with a phishing-only rationale. Inc.\ is the percentage of such inconsistent instances in the evaluation split. We validate the parser with a separate five-annotator audit; details and performance statistics are provided in Appendix~\ref{appendix:parser_audit}. The full cue-pattern material is in the supplementary repository 
(Appendix~\ref{app:repro}).

\paragraph{Evaluation aggregation.}
Table~\ref{tab:headline} reports Macro-F1 separately for ID/OOD and Test/Challenging settings. For each modality, we compute scenario-balanced averages so that each scenario contributes equally regardless of its sample size:
\[
M_{m,d,s}
=
\frac{1}{K_m}
\sum_{k=1}^{K_m}
M_{m,k,d,s},
\]
where \(m \in \{\mathrm{SMS},\mathrm{Voice}\}\), \(d \in \{\mathrm{ID},\mathrm{OOD}\}\), \(s \in \{\mathrm{Test},\mathrm{Challenging}\}\), and \(K_m\) is the number of scenarios for modality \(m\). For ID settings, \(k\) indexes scenario-specific subsets of the random split; for OOD settings, \(k\) indexes held-out scenario folds.

For the mechanism ablation in Table~\ref{tab:mechanism} and the \(\lambda\)-sweep deltas in Figure~\ref{fig:lambda_sweep}, we use condition-balanced averages:
\[
M_{\mathrm{cond}}
=
\frac{1}{8}
\sum_{m,d,s}
M_{m,d,s},
\]
where the sum ranges over two modalities, two domains, and two difficulty settings. This gives equal weight to each evaluation condition. We further report Hard F1 as the average Macro-F1 over OOD Challenging conditions across both modalities:
\[
F1_{\mathrm{hard}}
=
\frac{1}{2}
\left(
M_{\mathrm{SMS},\mathrm{OOD},\mathrm{Challenging}}
+
M_{\mathrm{Voice},\mathrm{OOD},\mathrm{Challenging}}
\right).
\]

\subsection{RQ1: Does ID Predict SL-OOD Robustness?}

\label{sec:exp:rq1}

Among trainable baseline families, ID Macro-F1 substantially overestimates OOD robustness (Table~\ref{tab:headline}): feature-based, neural, and encoder PLM models often achieve high ID performance, yet several degrade by large margins under scenario-level holdout. Specifically, KoBERT drops from \(77.77\) ID Macro-F1 to \(45.57\) OOD Macro-F1 on Voice Test, a \(32.20\)-point gap. TF-IDF + SVM shows a similar pattern on both modalities, dropping from \(97.29\) to \(74.92\) on SMS Test and from \(99.13\) to \(80.33\) on Voice Test. 

Figure~\ref{fig:id_ood_gap} visualizes this gap directly for trainable model configurations. We exclude proprietary LLMs from this figure because they are not fit on the ID training distribution, making their ID--OOD gap less directly interpretable as robustness degradation. Models with similarly high ID Macro-F1 can experience substantially different ID--OOD gaps, showing that random-split rank does not reliably estimate scenario-level robustness loss. ECoG variants cluster in the high-ID, low-gap region across panels, while several baseline families remain vulnerable to large OOD gaps.

\subsection{RQ2: Which Model Families Generalize?}
\label{sec:exp:rq2}

By model family (Table~\ref{tab:headline}), \emph{feature-based} TF-IDF + SVM achieves strong ID performance but degrades sharply under scenario shift, consistent with reliance on scenario-specific lexical cues. \emph{Neural} baselines improve OOD performance but remain sensitive to challenging splits (\(58.19\)--\(70.32\) on SMS Challenging OOD). \emph{Encoder PLMs} narrow the SMS ID--OOD gap but transfer poorly to Voice.

\emph{Decoder-based} models behave qualitatively differently. KULLM-5B (QLoRA) and HCX-0.5B ST-SFT attain higher OOD Macro-F1 than encoder baselines on both modalities, with HCX-0.5B ST-SFT reaching \(84.98\) on SMS Challenging OOD. \emph{Rationale-only} variants transfer inconsistently: HCX-0.5B Rat.-only stays competitive on SMS Test OOD but drops on SMS Challenging OOD, while Qwen3-0.6B Rat.-only degrades sharply on Voice OOD (down to \(62.84\) on Test and \(59.90\) on Challenging). Proprietary LLMs are a strong reference point without task-specific fine-tuning, but ECoG wins three of four OOD columns and is the most consistent trainable family across modalities, difficulty settings, and backbones; GPT-5.4 (\(90.42\)) remains strongest on Voice Challenging OOD.

\begin{table}[t!]
\centering
\caption{ECoG performance across all decoder backbone sizes, reported as scenario-balanced Macro-F1. HCX-0.5B and Qwen3-0.6B rows duplicate the corresponding ECoG entries in Table~\ref{tab:headline} for direct scale comparison.}

\label{tab:scale}
\small
\renewcommand{\arraystretch}{1}
\setlength{\tabcolsep}{2.8pt}

\begin{tabular}{crrrrrrrr}
\toprule
\multirow{4}{*}{\textbf{Model}}
& \multicolumn{4}{c}{\textbf{SMS}}
& \multicolumn{4}{c}{\textbf{Voice}} \\
\cmidrule(lr){2-5}\cmidrule(lr){6-9}
& \multicolumn{2}{c}{\textbf{Test}}
& \multicolumn{2}{c}{\textbf{Challenging}}
& \multicolumn{2}{c}{\textbf{Test}}
& \multicolumn{2}{c}{\textbf{Challenging}} \\
\cmidrule(lr){2-3}\cmidrule(lr){4-5}\cmidrule(lr){6-7}\cmidrule(lr){8-9}
& \multicolumn{1}{c}{ID} & OOD & ID & OOD & ID & OOD & ID & OOD \\
\midrule
HCX-0.5B
& {99.65} & {94.97} & {96.80} & {92.71} & {100.00} & {94.56} & {97.36} & {88.98} \\
HCX-1.5B
& {99.62} & 95.04 & 96.18 & {93.01}
& 99.66 & {95.15} & 97.47 & 87.44 \\
Qwen3-0.6B
& {99.40} & {94.93} & {97.88} & {89.75} & {99.66} & {92.05} & {95.66} & 84.78 \\
Qwen3-1.7B
& 99.46 & {95.80} & {96.34} & 91.70
& {99.80} & 93.51 & {97.91} & {87.57} \\
\bottomrule
\end{tabular}
\end{table}

To isolate methodology effects from backbone scale, Table~\ref{tab:scale} reports all four backbone sizes. ECoG retains strong OOD performance at larger scale, with scale effects varying by modality and difficulty setting; the consistency-weight sensitivity in Figure~\ref{fig:lambda_sweep} appears at both scales.
 
\subsection{RQ3: Which Objectives Drive Gains?}
\label{sec:exp:ablation}
\begin{table}[t!]
\centering
\caption{Mechanism ablation on HCX-0.5B.
Mechanism ablation on HCX-0.5B; metrics are defined in Section~\ref{sec:exp:setup} (lower is better for Inc., higher for the rest).
Evi., Rat., and C denote evidence, rationale, and consistency objectives; \emph{+ Last} uses last-token pooling for the consistency head; \emph{Label-Last} places the label field last in the target sequence. Dashes mark metrics the method does not produce.
}
\label{tab:mechanism}
\small
\setlength{\tabcolsep}{1.45pt}
\renewcommand{\arraystretch}{1}
\begin{tabular}{@{\hspace{3pt}}>{\raggedright\arraybackslash}p{0.41\columnwidth}
C{0.100\columnwidth}
C{0.100\columnwidth}
C{0.095\columnwidth}
C{0.100\columnwidth}
C{0.125\columnwidth}@{}}
\toprule
\textbf{Method}
& \textbf{Cls}
& \textbf{Hard}
& \textbf{Inc.}
& \textbf{Span}
& {\scriptsize\textbf{BERTScore}} \\
\midrule

\multicolumn{6}{@{\hspace{3pt}}l}{\textbf{\textit{Basic Baselines}}} \\

\quad ST-SFT
& 91.60 & 83.30 & -- & -- & -- \\

\addlinespace[2pt]
\multicolumn{6}{@{\hspace{3pt}}l}{\textbf{\textit{Pipeline Baseline}}} \\
\quad 2-Stage SFT
& 89.37 & 81.96 & -- & -- & -- \\

\addlinespace[2pt]
\multicolumn{6}{@{\hspace{3pt}}l}{\textbf{\textit{Objective Ablation}}} \\
\quad Label + Evi.
& 94.66 & 86.57 & -- & 14.67 & -- \\
\quad Label + Rat. w/o C \((\lambda=0)\)
& 88.58 & 77.33 & 10.70 & -- & 74.15 \\
\quad Label + Rat. + C \((\lambda=0.1)\)
& 94.38 & 86.29 & 4.91 & -- & 79.44 \\

\addlinespace[2pt]
\multicolumn{6}{@{\hspace{3pt}}l}{\textbf{\textit{Consistency Weight Sweep}}} \\
\quad Joint w/o C \((\lambda=0)\)
& 95.25 & 87.62 & 9.36 & 33.85 & 75.07 \\
\quad Joint + C \((\lambda=0.05)\)
& 95.06 & 88.38 & 4.46 & 40.73 & 78.50 \\
\quad \textbf{Joint + C \((\boldsymbol{\lambda=0.1})\)}
& 95.63 & 90.84 & 5.14 & \textbf{42.23} & \textbf{80.05} \\
\quad Joint + C \((\lambda=0.2)\)
& 94.77 & 87.35 & \textbf{3.49} & 41.39 & 78.63 \\

\addlinespace[2pt]
\multicolumn{6}{@{\hspace{3pt}}l}{\textbf{\textit{Pooling Ablation}}} \\
\quad Joint + C \((\lambda=0.1)\) + Last
& \textbf{96.31} & \textbf{91.62} & 7.31 & 37.56 & 73.64 \\

\addlinespace[2pt]
\multicolumn{6}{@{\hspace{3pt}}l}{\textbf{\textit{Last Label Ablation}}} \\
\quad Label-Last, w/o C \((\lambda=0)\)
& 81.16 	& 65.85 	& 3.78	& 38.96 	& 77.20 \\
\quad Label-Last, + C \((\lambda=0.1)\)
& 82.40 	& 68.06 	& 3.41	& 41.68 	& 79.67 \\

\bottomrule
\end{tabular}
\end{table}

\paragraph{Mechanism ablation.}
Adding evidence supervision to label-only ST-SFT (Table~\ref{tab:mechanism}) gives the largest gain among non-consistency variants, improving Cls F1 from \(91.60\) to \(94.66\) (\(+3.06\) points), which suggests that input-side span supervision is useful before rationale generation is introduced. In contrast, label-plus-rationale training without consistency lowers classification performance (\(88.58\) Cls F1 and \(77.33\) Hard F1) and produces a high inconsistency rate (\(10.70\%\) Inc.), indicating that free-text rationale supervision alone is not a reliable consistency signal. Adding the consistency loss largely removes this failure, improving Cls F1 to \(94.38\), Hard F1 to \(86.29\), and Inc.\ to \(4.91\%\). The two-stage SFT baseline also underperforms ST-SFT (\(89.37/81.96\) vs.\ \(91.60/83.30\) in Cls/Hard F1), suggesting that directly extending a label/evidence checkpoint to a longer generative target can disrupt rather than preserve discriminative behavior.

\begin{figure}[t!]
  \centering
  \includegraphics[width=\columnwidth]{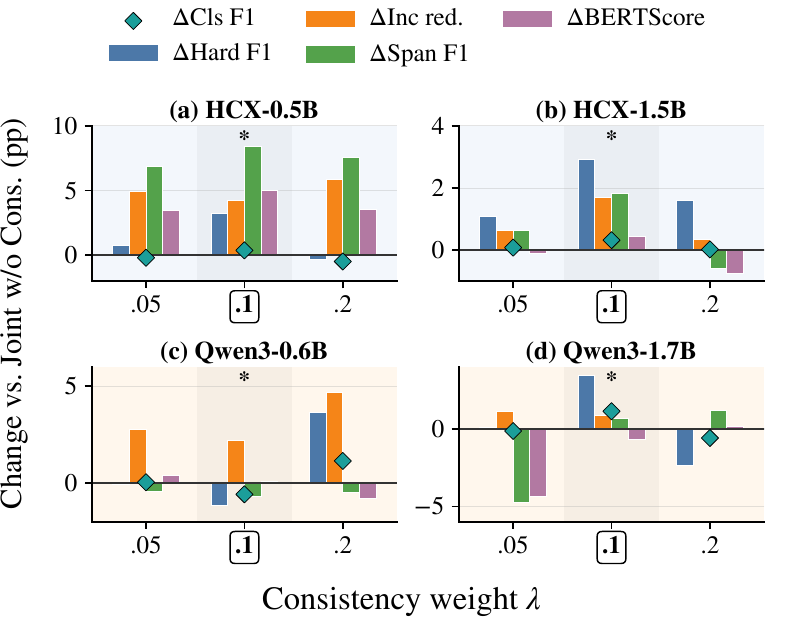}
\caption{Effect of the consistency weight $\lambda$ relative to Joint w/o C ($\lambda=0$). Bars show changes in Hard F1, Inc.\ reduction, Span F1, and BERTScore; diamonds show $\Delta$Cls F1. Boxed values indicate the selected ECoG configuration ($\lambda=0.1$).}
\Description{Four bar charts arranged in a two-by-two grid, one per decoder backbone: HCX-0.5B, HCX-1.5B, Qwen3-0.6B, and Qwen3-1.7B. Each chart plots the change in percentage points relative to training without the consistency loss, on the vertical axis, against consistency weights of 0.05, 0.1, and 0.2 on the horizontal axis. Grouped bars show change in Hard F1, inconsistency reduction, span F1, and BERTScore, and diamond markers show change in classification F1. The selected setting of 0.1 is boxed on the axis and marked with an asterisk. Inconsistency reduction is positive at every weight in all four panels, while the other metrics vary in sign and magnitude across backbones; the vertical scale differs by panel, spanning roughly 10 points for HCX-0.5B and 4 points for HCX-1.5B.}
\label{fig:lambda_sweep}
\end{figure}

\paragraph{Consistency-weight sensitivity.}
On HCX-0.5B, \(\lambda=0.2\) yields the lowest Inc.\ but lowers Cls F1 and Hard F1, whereas the default \(\lambda=0.1\) gives the best hard-case classification among the consistency-weight variants in Table~\ref{tab:mechanism}. Relative to Joint w/o C, \(\lambda=0.1\) improves Cls F1 by \(+0.38\) points and Hard F1 by \(+3.22\) points (averaged over the two OOD challenging conditions, of which the SMS folds are small; Table~\ref{tab:dataset_stats}); the corresponding Inc.\ reduction is reported alongside output-side metrics in Section~\ref{sec:exp:explanation}.

Figure~\ref{fig:lambda_sweep} extends the sweep across backbone families and model sizes. The default \(\lambda=0.1\) reduces Inc.\ relative to Joint w/o C in all four configurations and improves Hard F1 in three of four, although the magnitude varies by backbone. We therefore treat \(\lambda\) as a sensitivity analysis and use \(\lambda=0.1\) as the fixed headline setting rather than tuning it per backbone. Per-fold breakdowns (supplementary repository) confirm the same selectivity: consistency has small or mixed effects on Test OOD but more consistent gains on Challenging OOD, supporting its role as a hard-case regularizer rather than a uniform OOD accuracy booster.

Appendix~\ref{app:difficulty-tier} further shows that HCX-0.5B gains concentrate on harder human-judged tiers, whereas Qwen3-0.6B remains nearly flat, consistent with the backbone-dependent response in Figure~\ref{fig:lambda_sweep}.

\paragraph{Target order ablation.}
The label-last variant reduces Inc.\ but substantially degrades Cls and Hard F1 (Table~\ref{tab:mechanism}, Label-Last rows), suggesting that forcing the model to decide after a long evidence--rationale sequence is not beneficial for robust detection. We therefore retain the label-first order. Beyond accuracy, this order is also deployment-aligned: the detector emits the decision before generating supporting evidence and rationale, matching latency-sensitive settings where an early warning can be acted on before a fuller explanation is inspected, though we do not measure wall-clock latency.

\subsection{RQ4: Grounded and Consistent Outputs?}
\label{sec:exp:explanation}

Turning to generated-output behavior, ECoG (Joint + C, \(\lambda=0.1\)) improves over Joint w/o C (\(\lambda=0\)) on HCX-0.5B across all three output-side measures (Table~\ref{tab:mechanism}, Figure~\ref{fig:lambda_sweep}): Span F1 increases by \(+8.38\) points (\(33.85 \rightarrow 42.23\)), BERTScore F1, our rationale-similarity indicator, increases by \(+4.98\) points (\(75.07 \rightarrow 80.05\)), and Inc.\ decreases by \(4.22\) points (\(9.36 \rightarrow 5.14\)). BERTScore measures reference-rationale similarity, not logical sufficiency or causal faithfulness.

Across the four backbone configurations in Figure~\ref{fig:lambda_sweep}, the default consistency weight reduces Inc.\ relative to \(\lambda=0\) in all cases, although Span F1 and BERTScore are more backbone-dependent. Figure~\ref{fig:per-scenario-inconsistency} in Appendix~\ref{app:scenario-inconsistency} further shows that inconsistency reduction is generally positive but heterogeneous across held-out scenarios.

The pooling ablation shows a related trade-off between classification and generated-output quality. Last-token pooling yields higher Cls F1 (\(96.31\)) and Hard F1 (\(91.62\)) than mean pooling, but degrades output-side behavior: Inc.\ increases by \(2.17\) points, Span F1 decreases by \(4.67\) points, and BERTScore drops by \(6.41\) points relative to ECoG (Table~\ref{tab:mechanism}). We therefore retain mean pooling as the default.
\section{Discussion}
\label{sec:discussion}

\subsection{ID Accuracy Is Not Robustness}
\label{sec:disc:id-vs-ood}

\begin{figure}[t!]
  \centering
  \includegraphics[width=0.98\columnwidth]{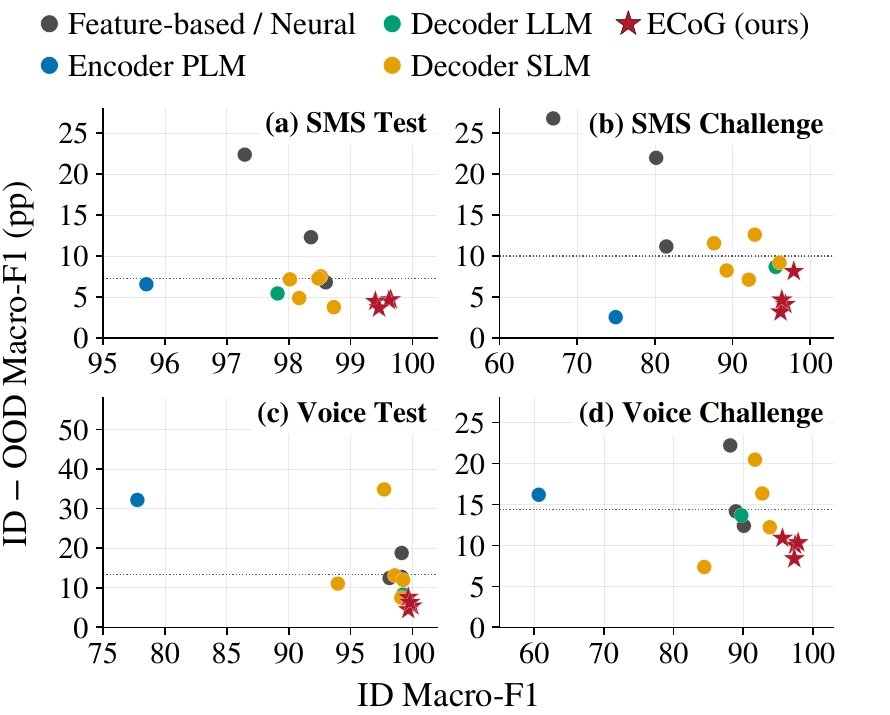}
\caption{ID--OOD robustness gap under scenario shift for trainable configurations, grouped by model family; stars denote ECoG. Lower values on the y-axis (ID minus OOD Macro-F1, in points) indicate stronger robustness. Dotted lines mark the per-panel average gap; panel (c) uses a wider y-range for the KoBERT Voice outlier.}
\Description{Four scatter plots arranged in a two-by-two grid, covering SMS Test, SMS Challenge, Voice Test, and Voice Challenge. Each plots in-distribution Macro-F1 on the horizontal axis against the in-distribution minus out-of-distribution Macro-F1 gap in percentage points on the vertical axis, so lower points indicate stronger robustness. Markers are colored by model family: feature-based and neural, encoder pretrained language models, decoder large language models, and decoder small language models, with ECoG variants shown as red stars. A dotted horizontal line marks the average gap in each panel. ECoG stars sit in the lower-right region of every panel, combining high in-distribution accuracy with small gaps, while several baselines show gaps above 20 percentage points; the vertical range of the Voice Test panel is wider to accommodate an encoder outlier near 32 points.}
  \label{fig:id_ood_gap}
\end{figure}

RQ1 and RQ2 show that random-split performance is a weak proxy for deployment robustness in phishing detection (Table~\ref{tab:headline}, Figure~\ref{fig:id_ood_gap}): comparable ID accuracy can hide large differences in scenario-level degradation, indicating that random splits reward recurring scenario-specific cues rather than intent-level generalization. Scenario-level holdout, especially with the challenging set, therefore gives a stricter deployment-oriented test.

\subsection{Evidence and Consistency Regularization}
\label{sec:disc:alignment}

RQ3 and RQ4 clarify the complementary roles of evidence supervision and rationale--label consistency regularization (Table~\ref{tab:mechanism}, Figure~\ref{fig:lambda_sweep}). Evidence supervision anchors training to input-side spans that support the decision, whereas the consistency objective encourages generated rationales to remain directionally consistent with the predicted label. This distinction matters because free-text rationale supervision alone can produce fluent rationales that support the wrong label, consistent with prior reports of unfaithful chain-of-thought rationalization~\cite{turpin2023language}. The consistency objective primarily affects hard-case and generated-output behavior: Cls F1 changes modestly while Hard F1, Span F1, and BERTScore improve and Inc.\ decreases (Table~\ref{tab:mechanism}). We therefore interpret the consistency branch as a hard-case training-time regularizer rather than a general-purpose OOD accuracy booster.

This interpretation does not establish causal faithfulness of the generated rationale. The consistency branch biases rationale-position representations toward label-discriminative information while leaving the inference-time decoding path unchanged, but it is not a post-hoc verifier of rationale sufficiency.

\section{Conclusion and Future Work}
\label{sec:conclusion}

We formulated Korean SMS and voice phishing detection as scenario-level OOD generalization and constructed a holdout protocol with a challenging set. We proposed ECoG, a decoder-based generative framework that combines evidence-span supervision with rationale--label consistency regularization. Across two backbone families and four model sizes, ECoG consistently reduces prediction--rationale inconsistency, with backbone-dependent gains in hard-case classification and evidence-span overlap. These results show that random-split evaluation can mislead deployment-oriented model selection, and that evidence-consistent training improves compact generative detection under social-engineering shift.

There are several future directions to improve ECoG.
Our evaluation focuses on Korean SMS and voice phishing, leaving other languages, communication channels, and temporal updates for future work. The voice setting relies on \textit{whisper-small}~\cite{radford2022whisper} ASR transcripts, so performance may inherit transcription errors. We will evaluate robustness to ASR noise and end-to-end audio--text inputs. Our explanation metrics also do not establish causal faithfulness. Rationale-position hidden states remain conditioned on preceding target tokens (Section~\ref{sec:method:cons}), and the rule-based Inc.\ parser may miss malformed or semantically inconsistent rationales outside its cue patterns. 

Extending the protocol to include adversarial within-scenario perturbations, temporal concept drift, and stronger causal-rationale tests would further strengthen deployment-oriented assessment.

\appendix
\section{Annotation Details}
\label{appendix:annotation}

\paragraph{Annotation procedure.}
For each instance, we use \textit{GPT-4o-mini} with temperature \(0\) to construct evidence-span and rationale annotations for training supervision. Given the input text \(x\) and gold label \(y\), the annotation prompt asks the model to copy evidence spans verbatim from \(x\) and generate a Korean natural-language rationale conditioned on the selected evidence. For phishing instances, evidence spans target functional cues such as impersonation, urgency, malicious-link inducement, credential solicitation, or payment requests. For benign instances, the evidence field is set to a fixed ``none'' marker unless the message contains an explicit benign-supporting cue.

\paragraph{Filtering and reproducibility.}
We discard evidence spans that are not exact substrings of the input, duplicate another span, or consist only of trivial formatting artifacts. 

\subsection{Annotation Quality and Privacy}
\label{sec:annotation_quality}

\paragraph{PII de-identification.}
Prior to model training and evaluation, we de-identify the dataset to remove personally identifiable information such as names, addresses, phone numbers, and account numbers. We use \textit{GPT-4o-mini} to detect and redact PII using a fixed redaction prompt. A manual audit of a randomly sampled 10\% subset found no remaining sensitive PII; the main failure mode was conservative over-redaction, affecting 0.874\% of spans, typically for non-identifying fields such as timestamps and dates.

\paragraph{Evidence--rationale audit.}
We assess annotation quality on \(1{,}916\) stratified-sampled instances. Each item received judgments from five independent annotators on a three-point ordinal scale (\emph{Pass} / \emph{Acceptable} / \emph{Invalid}). The majority-vote distribution is \(69.89\%\) Pass, \(26.98\%\) Acceptable, and \(3.13\%\) Invalid, corresponding to a \(96.87\%\) usable annotation rate. Inter-annotator agreement is Fleiss' \(\kappa = 0.8047\) and Krippendorff's nominal \(\alpha = 0.8047\), indicating substantial agreement under the Landis--Koch scale~\cite{measurement}. Unanimous agreement (\(5/5\) annotators) is reached on \(75.68\%\) of items, and the mean majority size is \(4.76/5\) (\(95.14\%\)). These results support using the filtered annotations as reference evidence and rationales for supervision and evaluation.

\section{Challenging Set Construction}
\label{appendix:challenging_set}

\paragraph{Overview.}
We construct a challenging evaluation set of ambiguous or lexically confusable instances that require intent-level discrimination. The challenging set is used only for evaluation and is never included in model training. It is intended as a stress test rather than an adversarial attack benchmark.

\paragraph{SMS challenging set.}
The SMS challenging set is constructed through error-driven mining followed by human-difficulty filtering. We train multiple encoder baselines under cross-validation and collect phishing instances misclassified by at least one baseline as a candidate pool of lexically confusable or decision-boundary cases. The mining baselines are DistilKoBERT~\cite{park2019distilkobert}, DistilBERT~\cite{sanh2020distilbertdistilledversionbert}, KoBERT~\cite{kobert_sktbrain_github}, and mBERT~\cite{pires2019multilingualmultilingualbert}. To construct paired benign cases, we retrieve lexically similar benign candidates using TF--IDF character \(n\)-gram similarity with length matching. Five blind annotators then label candidate instances without access to model predictions; instances are retained when at least three of five annotators disagree with the ground-truth label, or when one to two annotators disagree and the correct annotators' mean confidence is below \(0.5\) on a unit-interval confidence scale.

\paragraph{Voice challenging set.}
For voice phishing, we use institutionally verified victim-case transcripts as the phishing portion of the challenging set. For each phishing instance, we retrieve the most semantically similar benign transcript from the benign voice corpus as a paired negative example. This simpler procedure is used because the voice corpus is smaller and the positive challenging cases already come from institutionally verified cases.

\begin{figure}[t!]
  \centering
  \includegraphics[width=\columnwidth]{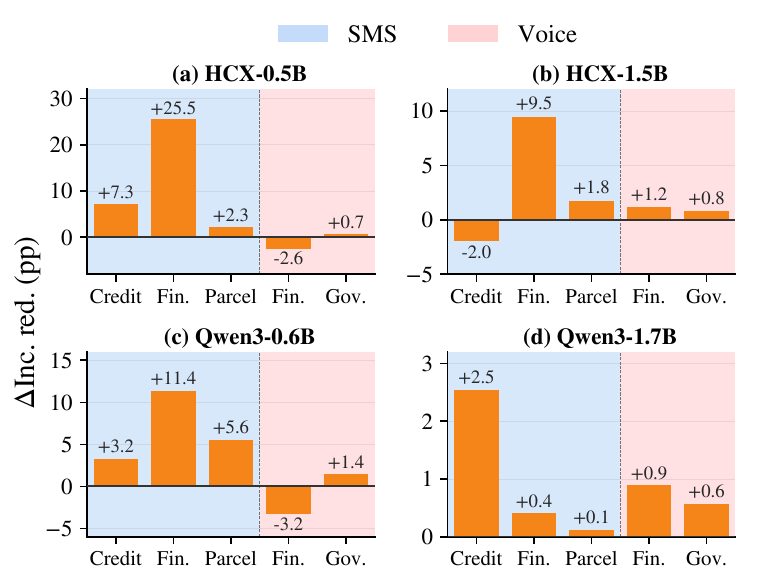}
\caption{Per-scenario reduction in prediction--rationale inconsistency by ECoG relative to Joint w/o C across backbones. Positive values indicate lower inconsistency under ECoG; negative values indicate increased inconsistency. Background colors denote SMS and voice scenarios.}
\Description{Four bar charts arranged in a two-by-two grid, one per decoder backbone: HCX-0.5B, HCX-1.5B, Qwen3-0.6B, and Qwen3-1.7B. Each shows the reduction in prediction-rationale inconsistency in percentage points achieved by ECoG relative to training without the consistency loss, across five held-out scenarios: Credit, Finance, and Parcel for SMS, and Finance and Government for Voice. Background shading separates the SMS and Voice scenario groups. Most bars are positive, with the largest reduction on SMS Finance for HCX-0.5B at 25.5 points, while a few settings are negative, including Voice Finance for HCX-0.5B and Qwen3-0.6B and SMS Credit for HCX-1.5B. The vertical scale differs by panel, ranging from about 30 points for HCX-0.5B down to about 3 points for Qwen3-1.7B.}
  \label{fig:per-scenario-inconsistency}
\end{figure}

\section{Training Configuration}
\label{app:config}

All results are reported from a single training run; we use \(\Delta\) values for sensitivity analyses (Figure~\ref{fig:lambda_sweep}, Table~\ref{tab:mechanism}) to mitigate single-seed noise.

We fix $\beta_{\mathrm{label}}=1$, $\beta_{\mathrm{gen}}=0.5$, $\beta_{\mathrm{cls}}=0.5$, and $\beta_{\mathrm{evi}}=1$ across all experiments. The prompt-side classification loss applies positive-class weighting $w_+=N_0/N_1$ (range 0.770--1.335 across training splits). The consistency weight $\lambda$ is swept in $\{0.0, 0.05, 0.1, 0.2\}$, with $\lambda=0.1$ used as the default. Per-split weight values and full hyperparameter settings are provided in the supplementary repository (Appendix~\ref{app:repro}).

\section{Parser--Human Consistency Audit}
\label{appendix:parser_audit}

We validate the Inc.\ parser with a separate blinded audit over 1{,}916 SMS and voice challenging cases. Five annotators are shown the input, model prediction, generated rationale, and generated evidence when available, but not the gold label, parser decision, model identity, or \(\lambda\) value. Each instance is labeled as \emph{Consistent}, \emph{Unclear/Mixed}, or \emph{Inconsistent}, where \emph{Inconsistent} denotes that the rationale primarily supports the opposite label or contradicts the predicted label.

Human majority vote identifies 120 inconsistent instances. The Inc.\ parser flags 113 of them and 6 additional cases judged by humans as Consistent or Unclear/Mixed, yielding 95.0\% precision, 94.2\% recall, 94.6\% F1, and 99.3\% overall parser--human agreement. The 13 disagreements are mostly borderline or mixed rationales, supporting Inc.\ as a conservative proxy for prediction--rationale direction mismatch rather than a measure of causal faithfulness.

\section{Per-Scenario Inconsistency}
\label{app:scenario-inconsistency}
Figure~\ref{fig:per-scenario-inconsistency} shows that ECoG generally reduces prediction--rationale inconsistency across held-out scenarios and backbone configurations, with the largest single reduction on SMS Finance for HCX-0.5B (\(+25.5\) points), and consistently positive but smaller reductions on the same scenario for the other backbones. The effect is nevertheless heterogeneous: several voice-finance settings show smaller or negative reductions, suggesting that consistency regularization improves output-level rationale--label consistency overall but does not eliminate scenario-specific variation.

\begin{table}[t!]
\centering
\caption{ECoG vs.\ Joint w/o C by human-difficulty tier on the retained SMS Challenging set. Scores are instance-level Macro-F1, not scenario-balanced averages; Table~\ref{tab:headline} reports scenario-balanced results. Difficulty tiers indicate how many of five blind annotators disagreed with the verified label during challenging-set construction.}
\label{tab:difficulty_tier}
\scriptsize
\renewcommand{\arraystretch}{1}
\setlength{\tabcolsep}{3pt}
\resizebox{\columnwidth}{!}{
\begin{tabular}{lrrrrrrr}
\toprule
\multirow{2.5}{*}{\textbf{Difficulty Tier}} & \multirow{2.5}{*}{\textbf{N}}
& \multicolumn{3}{c}{\textbf{HCX-0.5B}}
& \multicolumn{3}{c}{\textbf{Qwen3-0.6B}} \\
\cmidrule(lr){3-5}\cmidrule(lr){6-8}
& & w/o C & ECoG & \(\Delta\)Cls
& w/o C & ECoG & \(\Delta\)Cls \\
\midrule
\(3/5\) disagree & 82  & 95.11 & 96.31 & \(+1.20\) & 93.89 & 93.89 & \(+0.00\) \\
\(4/5\) disagree & 216 & 92.11 & 94.88 & \(+2.77\) & 93.96 & 93.93 & \(-0.03\) \\
\(5/5\) disagree & 257 & 86.77 & 91.42 & \(+4.65\) & 86.38 & 86.75 & \(+0.37\) \\
\midrule
All challenging & 555 & 90.08 & 93.49 & \(+3.41\) & 90.44 & 90.60 & \(+0.16\) \\
\bottomrule
\end{tabular}
}
\end{table}
\paragraph{Human-difficulty validation.}
\label{app:difficulty-tier}
We further validate the SMS challenging set by stratifying retained instances by blind human difficulty. The verified label is retained as ground truth, and human judgments are used only as a difficulty signal. The tiers indicate how many of five blind annotators disagreed with the verified label during construction. Table~\ref{tab:difficulty_tier} reports instance-level Macro-F1, rather than scenario-balanced Macro-F1 as in Table~\ref{tab:headline}. For HCX-0.5B, where consistency regularization yields strong gains overall, the ECoG improvement over Joint w/o C grows monotonically with human-judged difficulty. In contrast, Qwen3-0.6B shows a nearly flat pattern, consistent with the backbone-dependent sensitivity observed in Figure~\ref{fig:lambda_sweep}. This analysis supports the interpretation that consistency regularization acts as a hard-case regularizer when it provides meaningful gains for a given backbone.

\section{Reproducibility and Supplementary Materials}
\label{app:repro}
The supplementary repository includes:

\begin{itemize}
  \item De-identified SMS and voice datasets with scenario labels and challenging-set splits.
  \item Annotation prompts (evidence/rationale, PII redaction).
  \item Per-split class weights, optimization hyperparameters, and training scripts.
  \item Parser source code containing the cue-pattern rules used in output parsing and normalization.
  \item Per-fold OOD Macro-F1 breakdowns.
  \item Model configurations and evaluation scripts.
\end{itemize}

\begin{acks}
We would like to thank the anonymous reviewers for their helpful questions and comments.
This work was partly supported by Institute of Information \& communications Technology Planning \& Evaluation(IITP) grant funded by the Korea government(MSIT)
(RS-2019-II190421, Artificial Intelligence Graduate School Program (Sungkyunkwan University) \& 
RS-2025-02263169, Detection and Prediction of Emerging and Undiscovered Voice Phishing \&
RS-2024-00398115, Research on the reliability and coherence of outcomes produced by Generative AI)
 and the Ministry of Education of the Republic of Korea and the National Research Foundation of Korea (NRF-RS-2025-00523385).
\end{acks}

\clearpage

\section*{GenAI Usage Disclosure}
GPT-4o-mini~\cite{openaigpt4omini} was used to construct evidence-span and rationale annotations and to redact personally identifiable information, as described in Section~\ref{sec:data:construction}; no other generative AI was used for data generation, augmentation, or filtering, and challenging-set construction relied on encoder baselines and human annotators only. Generative AI assistants were used for plotting scripts, debugging, and refactoring, while the training, evaluation, and metric code was written by the authors; no reported experimental result was produced or modified by a generative AI tool. For the manuscript, generative AI tools were used for language editing and LaTeX formatting, but no section was drafted verbatim and reference collection was done by the authors. Generative AI also assisted in scoping the research problem, whereas the method design, baseline selection, and interpretation of results were carried out by the authors.

\bibliographystyle{ACM-Reference-Format}
\bibliography{references}

\end{document}